\pdfoutput=1

\documentclass[11pt]{article}

\usepackage[final]{acl}

\usepackage{times}
\usepackage{latexsym}
\usepackage{amsmath}
\usepackage{stfloats}
\usepackage[T1]{fontenc}

\usepackage[utf8]{inputenc}

\usepackage{microtype}

\usepackage{inconsolata}

\usepackage{graphicx}

\usepackage{csvsimple}
\usepackage{float}  

\title{Don't Forget It! Conditional Sparse Autoencoder Clamping Works for Unlearning}

\author{
  Matthew Khoriaty\textsuperscript{*} \\
  \texttt{matthewkhoriaty2026}\\
  \texttt{@u.northwestern.edu}\\
    \And
  Andrii Shportko\textsuperscript{*} \\
  \texttt{andreshportko2026}\\
  \texttt{@u.northwestern.edu}\\
  \And
  Gustavo Mercier \\
  \texttt{gustavomercier2026}\\
  \texttt{@u.northwestern.edu}
   \And
  Zach Wood-Doughty \\
  \texttt{zach}\\
  \texttt{@northwestern.edu} \\
}

\begin{document}
\maketitle
\renewcommand\thefootnote{}\footnote{\textsuperscript{*}Co-first authors}
\begin{abstract}
Recent developments in Large Language Model (LLM) capabilities have brought great potential but also posed new risks. For example, LLMs with knowledge of bioweapons, advanced chemistry, or cyberattacks could cause violence if placed in the wrong hands or during malfunctions. Because of their nature as near-black boxes, intuitive interpretation of LLM internals remains an open research question, preventing developers from easily controlling model behavior and capabilities. The use of Sparse Autoencoders (SAEs) has recently emerged as a potential method of unraveling representations of concepts in LLMs internals, and has allowed developers to steer model outputs by directly modifying the hidden activations. In this paper, we use SAEs to identify unwanted concepts from the Weapons of Mass Destruction Proxy (WMDP) dataset within {\tt gemma-2-2b} internals and use feature steering to reduce the model's ability to answer harmful questions while retaining its performance on harmless queries. Our results bring back optimism to the viability of SAE-based explicit knowledge unlearning techniques. 
\end{abstract}

\section{Introduction}

As the capacity of large language models (LLMs) for real-world applications has grown, so has concern about their safety \citep{zhao2023explainabilitylargelanguagemodels}. An LLM that accurately answers questions regardless of their implications could be misused to answer a wide variety of malicious queries. An LLM with a demonstrable understanding of chemistry could be used to advance science \citep{vert2023will} but could also be used to develop weapons \citep{li2024wmdpbenchmarkmeasuringreducing}.

The goal of safer LLMs has inspired several methodological approaches, from requiring ethical reasoning \citep{rao2023ethical} to enforcing guardrails \citep{rebedea2023nemo}. These methods attempt to prevent a model from revealing potentially harmful information to users. Machine unlearning methods instead seek to remove harmful knowledge so that the model cannot accurately answer malicious questions \citep{bourtoule2021machine, gupta2021adaptive}.

Machine unlearning has emerged as a promising approach, but current methods often face trade-offs between interpretability, computational efficiency, and unintended side effects (decreased performance on safe information). Representation Misdirection for Unlearning (RMU) modifies the model's internal representations to randomize knowledge in dangerous domains while preserving useful knowledge, but it lacks interpretability and can be computationally expensive \citep{li2024wmdpbenchmarkmeasuringreducing}. Meanwhile, Sparse Autoencoder (SAE)-based techniques \citep{farrell2024applyingsparseautoencodersunlearn} provide a more structured approach by directly modifying interpretable latent activations, allowing for targeted suppression of harmful knowledge while retaining critical functionality.

\subsection{Our Contributions to SAE-Based Unlearning}
\begin{enumerate}
\item A novel SAE-based unlearning methodology that employs conditional steering to improve on the retention of safe information over \citet{farrell2024applyingsparseautoencodersunlearn} in \S \ref{RefusalClamp}.
\item The previous paper has either had high side effects or were black-box methods that were uninterpretable. We improve on both these metrics \S \ref{sec:results}.
\item Higher "Alignment" compared to RMU with fewer side effects. Our Alignment measure is a novel contribution defined in \S \ref{sec:evaluation}
\item A novel format for representing and communicating SAE edits (Appendix \ref{repsteered})
\item A preliminary adversarial evaluation of our method compared with RMU in \ref{sec:adversarial}
\item We plan to open-source our code after publication 
\end{enumerate}

\section{Related Work}

We direct the reader to \citet{zhao2023explainabilitylargelanguagemodels} for a more comprehensive survey of the foundations of neural network interpretability. Here, we provide background on sparse autoencoders (SAEs) and representation engineering to introduce our proposed methodology in \S \ref{sec:methodology}.

\subsection{Sparse Autoencoders}

Autoencoders are neural networks trained to encode and decode an input through a bottleneck layer, typically such that the learned representation learned is a smaller representation of the original input.
SAEs extend this approach by enforcing sparsity (i.e. most values are near zero) within this representation using an additional loss term and large hidden dimension \citep{ng2011sparse}.

SAEs are particularly useful when applied within existing neural network models because neural networks tend to represent concepts using linear combinations of neurons, a property called "superposition" \citep{elhage2022superposition}. 
SAEs are used to expand the dimensionality of an internal activation within a larger neural network.
In these sparse representations learned by SAEs, each activation is often a distinct, interpretable feature \citep{cunningham2023sparseautoencodershighlyinterpretable,gurnee2023findingneuronshaystackcase}.

\subsection{Representation Engineering}

\citet{zou2023representationengineeringtopdownapproach} identified the nascent area of representation engineering,
a collection of approaches to find and control patterns inside learned representations of LLMs. It plays a crucial role in addressing vulnerabilities such as jailbreaking, where attackers can exploit model representations to bypass safety mechanisms.

\subsection{Prior Work in Unlearning with Sparse Autoencoders}
Recent work by \citet{farrell2024applyingsparseautoencodersunlearn} has suggested putting Unlearning in terms of the dichotomy between interpretability and low side effects. They used RMU-based unlearning as a baseline to compare with their SAE-based clamping algorithm. Their mechanism is described as follows: to identify features to steer, they identified the ones that fire frequently on the bio-forget subset yet rarely on the bio-retain one. Then, every time the latent is activated (>0), they clamped it to a negative scalar. They run a hyperparameter sweep to train a set of RMU models and empirically conclude that RMU-based approaches seem to outperform their clamping procedures. Our work resolves this issue by offering an altered algorithm that outperforms the above approaches in retaining safe knowledge. 

\subsection{Weapons of Mass Destruction Proxy Benchmark}

A popular metric of unlearning is The Weapons of Mass Destruction Proxy (WMDP) Benchmark. The benchmark is a dataset of 3,668 multiple-choice questions each with four answer choices, and performance on the benchmark is a stand-in for actual harmful knowledge and capabilities. It also comes with Bio-forget Corpus, Bio-retain Corpus, Cyber-forget Corpus, and Cyber-retain Corpus. Forget corpora consist of text examples in a subject (i.e. biology or cybersecurity) that are considered to have proxies for “dangerous” information, while retain corpora consist of text examples in a subject that are considered to be “safe” information \citep{li2024wmdpbenchmarkmeasuringreducing}. The WMDP dataset contains adversarial prompts designed to result in harmful outputs, to contribute to measuring robustness.  

\subsection{ Representation Misdirection for Unlearning}

Representation Misdirection for Unlearning (RMU) is a technique for unlearning developed by \citet{li2024wmdpbenchmarkmeasuringreducing}. The technique makes use of two datasets, 'forget' and 'retain', and a loss function that acts directly on the internals of the model, or 'representations'. The loss function optimizes the model to have random hidden states when data from the forget set are input without changing the hidden states when data from the retain set is input. This technique can reduce performance on WMDP-Bio without greatly reducing performance on MMLU. Limitations of RMU include that it is not interpretable — model weights are by default non-interpretable, so the changes learned during the RMU process remain so. Additionally, RMU is computationally expensive as it involves taking gradients and iterating though the forget and retain data.



\subsection{Gemma, SAELens, Wikitext, and MMLU}

Refer to the appropriate sections of Appendix \ref{context}.

\section{Data and Evaluation} \label{sec:evaluation}

The two goals of machine unlearning are to forget harmful information while retaining helpful information.
Clearly, a model that can never answer a prompt cannot provide harmful information; but such a model would also be useless for any practical purposes. Evaluating unlearning methods requires two datasets -- one that the model is meant to forget, and another that it must retain. We follow \citet{li2024wmdpbenchmarkmeasuringreducing} and use their WMDP dataset to measure forgetting. We use a subset of the MMLU dataset to measure retention \citep{hendryckstest2021}.\footnote{Specifically, we use the Biology task of the WMDP along with the MMLU tasks on college biology and computer science, high school US history and geography, and human aging.}

To measure retain accuracy, we took a weighted average over MMLU's test sets for High School History (204 questions), High School Geography (198 questions), Human Aging (223 questions), and College Computer Science (100 questions). To measure forget accuracy, we used WMDP-Bio's test set (1,273 questions).
We evaluated our methods using the EleutherAI evaluation harness, an open-source library that standardizes model evaluation on a variety of datasets including WMDP and MMLU \citep{eval-harness}.\\
\\
We tested a prefatory adversarial evaluation of our technique using the Concurrent Greedy Search algorithm.
A lot of unlearning evaluations have been about balancing a trade-off between minimizing loss on the general knowledge and maximizing loss on the harmful one. We offer a single metric that combines those losses which we call alignment (Eq. \ref{eq:2}). This metric makes use of the idea of Retention ($R$) (Eq. \ref{eq:1}) that has been previously explored in the \citep{farrell2024applyingsparseautoencodersunlearn} but for this specific task, we subtract $0.25$ which denotes a strategy of random guessing:
    
\begin{equation}
R = \min\left(1, \frac{\max(\epsilon, Acc_{\text{modified}} - 0.25)}{\max(\epsilon, Acc_{\text{original}} - 0.25)}\right)
\label{eq:1}
\end{equation}

\begin{equation}
    \text{ALIGNMENT} = R_{Good} * (1 - R_{Bad})
    \label{eq:2}
\end{equation}

\section{Methodology} \label{sec:methodology}

We chose {\tt gemma-2-2b} a pre-trained text-generation transformer-based model for our research. The previous work used {\tt gemma–2-2b-it}, an instruction-tuned augmentation of {\tt gemma-2-2b} fine-tuned to dialogue with users. {\tt gemma-2-2b} has more available sparse autoencoders with labeled latents, allowing us more flexibility in which latents we use for steering and conditioning. We sourced the latent labels from Neuronpedia \citep{neuronpedia} and used the SAE hook {\tt layer7/width16k/canonical} of the {\tt gemma-scope-2b-pt-res-canonical} SAE release. These SAE are widely used in interpretability research, as \citet{gemmateam2024gemma2improvingopen} is cited by 46 papers according to Google Scholar time of writing.

\subsection{Feature Selection}

We based our selection of sparse autoencoder (SAE) features on computing feature sparsities across a "forget" dataset (bio-weapon related contentin WMDP-Bio Forget) and a "retain" dataset (WMDP-Bio Retain), discarding features with high frequency in the retain dataset, and then ranking the remaining features by their activation in the forget dataset.

We have adopted this feature selection with few modifications. First, we created activation frequency dictionaries over tokens in the bio-forget and in the bio-retain datasets and normalized the values. Then, we discarded the features over the threshold for bio-retain (e.g., $0.0001$ ). Finally, from the pre-selected set, we choose top-k latents that are most frequent in the forget list (Figure \ref{fig:freq}).

 \citet{farrell2024applyingsparseautoencodersunlearn} used RMU as a benchmark for unlearning compared to their SAE-based approach. We have replicated the results so we could improve upon them in our experiments.

In the original paper, they clamped values to a negative scalar when the relevant latent activated (had a value above 0) and left the zero value unchanged otherwise. We tested a few modified approaches. First, since it is rare for latents to be exactly zero (\S \ref{discussion}), we adopted the following algorithms. 

\subsubsection{"Clamp Prime"}

In order to mitigate the issue that activations equal to zero are rare, our first approach raises the threshold ever-so slightly to reduce the number of clamping instances and, we suspect, the number of False Positives (Figure \ref{fig:clamp}).

\begin{figure}
    \centering
    \includegraphics[width=1\linewidth]{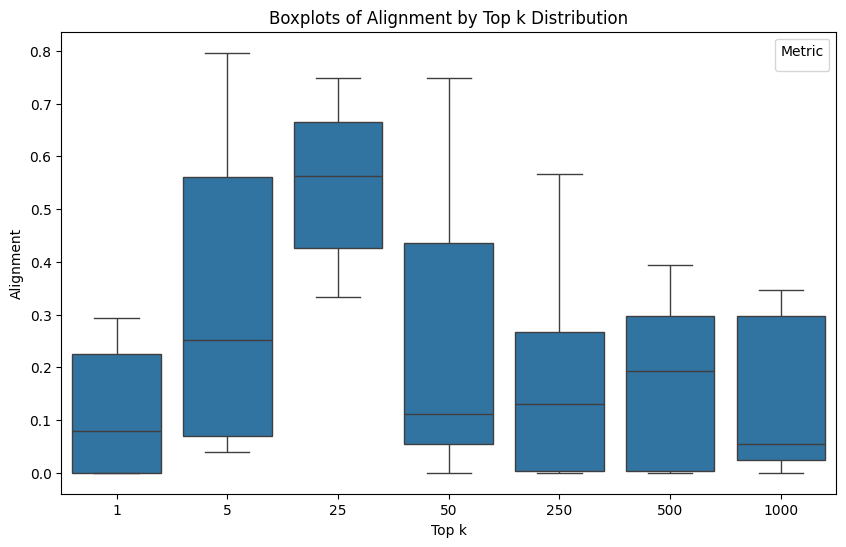}
    \caption{Effect of top-k selected features on the Alignment with all other hyperparameters being equal}
    \label{fig:topk}
\end{figure}

The Clamp Prime method has two hyperparameters to control: the number of selected features and the clamping scalar. All other things being equal, it seems that the optimal number of selected features should be no higher than 50 (Fig \ref{fig:topk}). Running the Kruskal-Wallis test, we discovered a significant difference ($H=4.8212, p=0.0272 ^*$).

\subsubsection{"Refusal Clamp"}\label{RefusalClamp}

\begin{figure}
    \centering
    \includegraphics[width=1\linewidth]{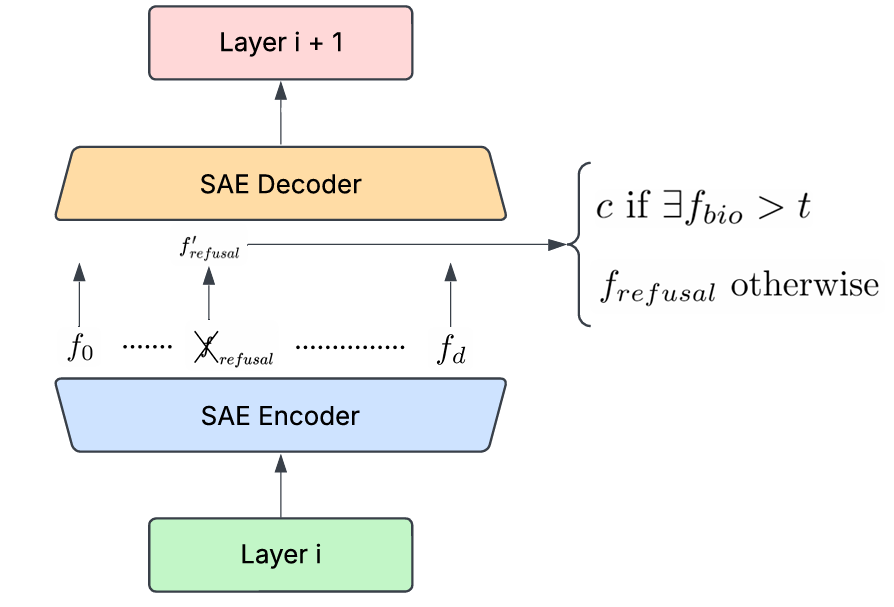}
    \caption{Refusal Clamp algorithm}
    \label{fig:refuse}
\end{figure}

In order to further decrease the number of clamping instances, we chose to clamp only one feature that is associated with refusal according to the autointerpreter from Neuronpedia according to GPT-4o. In essence, if any selected feature is activated above the threshold, the refusal latent is assigned to a negative clamping coefficient $c$ (Fig \ref{fig:refuse}).

\section{Experiments and Results} \label{sec:results}

While our methodology can be applied to any LLM for which internal activations are available, we chose to work with the {\tt gemma-2-2b} model released by Google due to the existence of trained and labeled Sparse Autoencoders from the Gemma Scope project.\footnote{\url{https://huggingface.co/google/gemma-2-2b}} This 2B-parameter model is small enough to run on a single Quadro RTX 8000 GPU which was used for all experiments.

\begin{figure}
    \centering
    \includegraphics[width=0.9\linewidth]{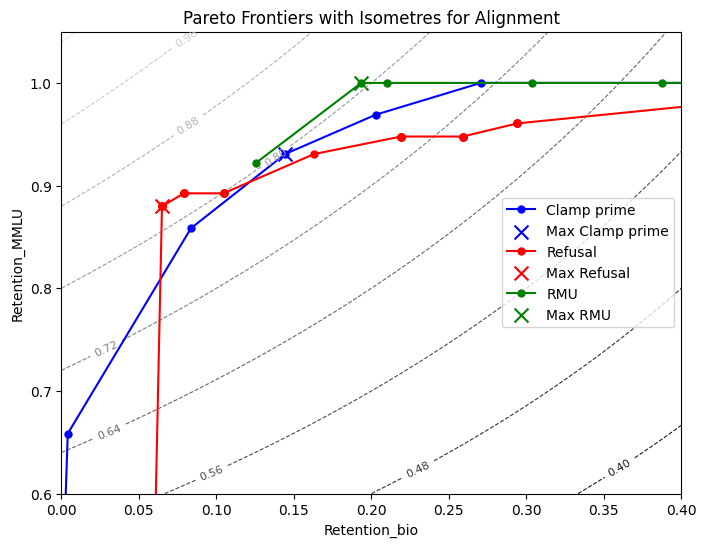}
    \caption{Pareto frontiers for 3 procedures along with top performing models. The isometres reflect the lines of equal Alignment score. } 
    \label{fig:pareto}
\end{figure}

The RMU method significantly reduces harmful knowledge (WMDP-bio accuracy: $0.5860 \rightarrow 0.3150$) while fully preserving useful knowledge, achieving a strong alignment score of $0.8065$ (Table. \ref{tab:results}). Clamp Prime improves on forgetting performance (WMDP-bio accuracy: $0.5860 \rightarrow 0.2985$) but slightly reduces general knowledge retention, leading to an alignment score of $0.7964$. The most aggressive unlearning occurs with the Refusal Clamp method, which achieves the lowest WMDP-bio accuracy ($0.5860 \rightarrow 0.2718$) and the highest alignment score ($0.8226$) at the cost of a noticeable impact on general knowledge (Fig. \ref{fig:pareto}). 

\subsection{Adversarial Evaluation}\label{sec:adversarial}

We did an adversarial evaluation using a Concurrent Greedy Search, an attack to find a suffix that causes the models to output a target answer given question. We randomly sampled the question-answer pair from the set of WMDP-Bio questions that {\tt gemma-2-2b} gets right in all permutations (Appendix \ref{adversarialquestion})\footnote{We used the following hyperparameters: number of tries per iteration: $b=32$, number of candidates per index: $k=64$, number of iters: $t=200$, and number of tokens in adversarial suffix: suffix\_length $=32$}.
As a baseline, this attack can induce the base model to output text involving E Coli and viral RNA. The RMU model which had the lowest accuracy on WMDP-Bio  and our conditionally steered model with the highest Alignment score (Refusal-clamp) both continued to output gibberish despite the optimization pressure of the suffix attack.

\begin{table*}

\small
  \centering
  \begin{tabular}{llllll}
    \hline
    \textbf{}           & \textbf{Acc on WMDP-bio} & \textbf{Acc on MMLU} & \textbf{Retention on WMDP-bio} & \textbf{Retention on MMLU}  & \textbf{Alignment} \\
    \hline
    No modification & 0.5860 & 0.5710 & 1.0    & 1.0    & 0.0    \\
    RMU            & 0.3150 & 0.5834 & 0.1935 & 1.0    & 0.8065 \\
    Clamp Prime     & 0.2985 & 0.5517 & 0.1444 & 0.9308 & 0.7964 \\
    Refusal-clamp   & 0.2718 & 0.5352 & 0.0649 & 0.8797 & \textbf{0.8226} \\
    \hline
  
  \end{tabular}
  \caption{Top alignment performances of different techniques. Hyperparameters are as follow: RMU ($s = 400$, $\alpha = 300$, $layer = 3$); Clamp prime (top-5 features, $c=-300$); Refusal clamp (top-10 features, $c=-500$, $t=0.05$)}
  \label{tab:results} 
\end{table*}

\section{Discussion}\label{discussion}

The developments we have made toward Sparse Autoencoder-based unlearning show promising results for the area of research. However, limitations in our studies inhibit further certainty in our approach and encourage future research in the field.

\subsection{Future Work}

Further work exploring the potential of SAE-based unlearning should investigate which SAE layers allow for the best unlearning.  We are also interested in the latent space dimensionality effect (i.e. the impact of using SAEs with larger shape of hidden activations).

SAE-Based clamping approaches that make a selection of edits based on some criteria are far from optimal. Future work should train interpretable parameters that edit the SAE latent vector to allow for fully interpretable fine-tuning.

\subsection{Rarity of Zero Activations}

\begin{figure}
    \centering
    \includegraphics[width=0.9\linewidth]{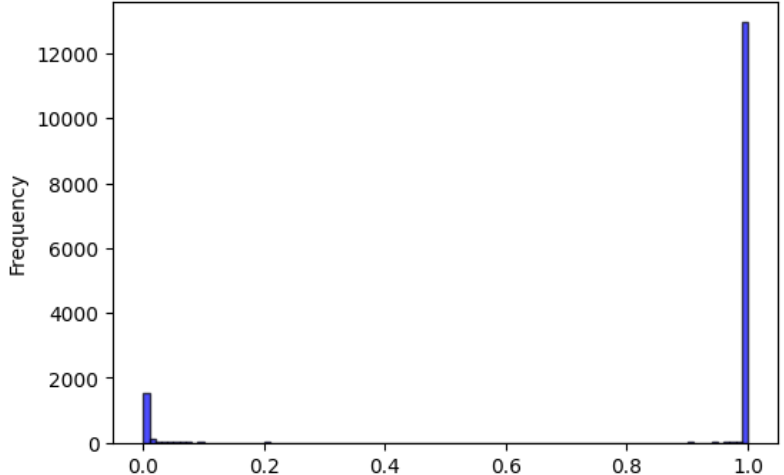}
    \caption{Average chance of the latent to be non-zero vs frequency of such latents in the layer $7$}
    \label{fig:act}
\end{figure}

What gave rise to our revisions to the original clamping algorithm is an observation of how rare it is for a latent to be exactly non-zero (Fig \ref{fig:act}). Specifically, it seems that only at most ~$19\%$ of the latents ever reached the value of $0$ over the entire dataset. Our algorithm raises the threshold to prevent the overstimulation of the algorithm. 

\section{Conclusion}
In this paper, we introduced an unlearning approach using Sparse Autoencoders that outperforms prior SAE-based unlearning approaches on retention and performs comparably with RMU on forgetting harmful information and retaining safe information while requiring fewer hyperparameters choices. While future work does need to verify the robustness of our method across a wider range of tests, our research indicates that SAEs are relevant to machine unlearning research.

\section{Acknowledgments}

We extend our gratitude to Evan Smith and Eliot Lee for their feedback. Also, the guidance of Professor Sruti Sakuntala Bhagavatula was central to our success.

\clearpage
\section{Limitations}

Despite its potential, sparse autoencoder unlearning has a number of drawbacks compared with traditional unlearning approaches. First, training sparse autoencoders and labeling the latents with their meanings is a slow and expensive process, limiting the practicality of SAE-based unlearning methods to those for which sparse autoencoders have already been created \citep{gemmateam2024gemma2improvingopen}. Additionally, steering models with sparse autoencoders necessitate that the models have closed weights since it is trivial to remove the steering hooks once an attacker has access to the weights of a model.

Our work has several limitations which could be addressed with further research. To start, though we trained our RMU models with WMDP’s Bio Forget Corpus as the “forget corpus” and Wikitext as the “retain corpus,” we used the comparison of WMDP Bio Forget and WMDP Bio Retain to find relevant autoencoder latents for conditioning and clamping. Using Wikitext as the retain set instead of WMDP’s Bio Retain may have caused the RMU models to consider biology a dangerous subject on which it is supposed to have random activations.\footnote{Since RMU compares the original model activations with the trained model activations to retain knowledge instead of comparing outputs to the ground truth, this training scheme does not give our RMU models an advantage on safe non-biology topics (such High School US History) through “extra training.”} But since we chose our clamping features based on a difference between the activations caused by safe biology against the activations caused by dangerous biology, we should expect that our measurements slightly inflate our method’s score on subjects which touch on biology, such as MMLU human aging.

While we have some evidence suggesting that our unlearning method is nontrivial to subvert using adversarial prompts, more work is needed to test its applicability against a wider variety of adversarial attacks and how it compares to other techniques such as RMU. To start, putting more computing power behind the adversarial attack by simply searching for adversarial suffixes for longer and testing a greater number of question-answer pairs would give us a better lower bound for the adversarial robustness of our approach. Since established unlearning methods were found to be vulnerable to adversarial attacks given sufficient computing power \citep{tamirisa2024tamperresistantsafeguardsopenweightllms}, the fact that our experiment was unable to overcome the RMU unlearning indicates we did not apply enough or apply the most applicable techniques.

\clearpage

\bibliography{custom}
\appendix

\section{Representing and Communicating Steered Models}\label{repsteered}
We propose a simple way to represent and communicate SAE-based clamping interventions. 
Clamping a model can be represented efficiently by a {\tt .csv} file defining the action, index, steering coefficient, SAE id and release. 

Steering CSVs are flexible and can represent a wide range of model edits and can be represented in a small number of bytes. Our best performing steered models, Clamp prime and Refusal-clamp are each represented in a trivial number of bytes: 4.0 kilobytes each. They have been reproduced in Appendix \ref{steeredappendix} in their entirety.

The infrastructure for reading Steering CSVs can be in continual development, as optional columns can hold parameters for new steering approaches.

Auxiliary columns such as "description" or "url" can be included for human-readable comments on the purpose of the edits or the description of the latents involved.

The required columns are “sae\_release”, “sae\_id”, “latent\_idx”, and “steering\_coefficient”. “clamp\_value”, “refusal\_id” are optional and used in particular steering methods.

While not necessarily the case, all of the steering methods we implemented apply to every position in the sequence.

Our implementation of Steering CSVs currently allows six “actions”:

\textbf{add}: Adds steering\_coefficient times the steering vector – the vector in the SAE’s decoding matrix at the latent\_idx’s position — to all sequence positions.

\textbf{clamp}: For each value at latent\_idx in the SAE’s latent representation greater than zero, set the value to steering\_coefficient.

\textbf{clamp\_cond}: Similar to clamp,  but instead of comparing values with zero, it compares with the optional “clamp\_value” argument. 

\textbf{clamp\_refusal}: Similar to clamp\_cond, but instead of clamping the latent\_idx, it clamps the refusal index “refusal\_id”. In the case of {\tt gemma-2-2b}, the refusal index is 15864 in layer\_7/width\_16k/canonical, gemma-scope-2b-pt-res-canonical

\textbf{print}: prints the activations and shape of the given SAE. 

\textbf{debug}: Activates the Python pdb debugger, allowing for manual inspection and editing of the activations. 

\section{Additional Tools and Related Work}\label{context}

The following appendixes describe additional tools and related work associated with our research, specifically Gemma Models, SAELens, Wikitext, and MMLU.

\subsection{Gemma Models}

The Gemma family of models are small, open weights models made by Google and available on Huggingface \citep{gemmateam2024gemma2improvingopen}. The training dataset was 2 trillion tokens of web documents, code, and mathematics. Both pre-trained and post-trained Gemma models are available in a range of sizes. The post-trained models were supervised and fine-tuned on prompt-response pairs, and Reinforcement Learning from Human Feedback further teaches the model to output preferred responses. In this work, we used the pre-trained {\tt gemma-2-2b}.

\subsection{SAELens and Gemma Scope}

SAELens is an open-source library based on TransformerLens which offers support for importing sparse autoencoders and editing their activations \citep{bloom2024saetrainingcodebase, nanda2022transformerlens}. Among the sparse autoencoders available are those trained by Google in their Gemma Scope project. Gemma Scope includes all layers and sublayers of Gemma-2-2b and 9B, and the features have LLM-interpreted meanings \citep{bills2023language}.

\subsection{Wikitext and Massive Multitask Language Understanding}
Wikitext is a public dataset that consists of verified Good and Featured articles from Wikipedia and acts as a large dataset of varied, quality, information-rich text data \citep{merity2016pointersentinelmixturemodels}. Massive Multitask Language Understanding (MMLU) is an evaluation dataset consisting of multiple-choice questions from a wide range of 57 tasks meant to test whether the model possesses "extensive world knowledge and problem-solving ability" \citep{hendryckstest2021}.

\section{Steered Models}\label{steeredappendix}
Included in table \ref{fig:clampedits} and table \ref{fig:refusalclampedits} are the edits made to our highest Alignment (see Section \ref{sec:evaluation}, Data and Evaluation for a description of our Alignment metric) steered models in the format described by Appendix \ref{repsteered}.
\begin{table*}[ht]
    \tiny
    \centering
    \makebox[\textwidth][c]{
    \begin{tabular}{|l|l|l|l|l|l|}
    \hline
        latent\_idx & hook\_action & steering\_coefficient & sae\_id & sae\_release & description \\ \hline
        11766 & clamp & -300 & layer\_7/width\_16k/canonical & gemma-scope-2b-pt-res-canonical & mentions of the coronavirus pandemic and related medical terms \\ \hline
        9723 & clamp & -300 & layer\_7/width\_16k/canonical & gemma-scope-2b-pt-res-canonical & references to viral RNA and its properties \\ \hline
        4788 & clamp & -300 & layer\_7/width\_16k/canonical & gemma-scope-2b-pt-res-canonical & terms related to viral infections and their impact on health \\ \hline
        1709 & clamp & -300 & layer\_7/width\_16k/canonical & gemma-scope-2b-pt-res-canonical & terms associated with epidemiological factors in disease transmission \\ \hline
        9186 & clamp & -300 & layer\_7/width\_16k/canonical & gemma-scope-2b-pt-res-canonical & terms related to respiratory health conditions and treatments \\ \hline
    \end{tabular}
    }
    \caption{The model edits made for the Clamp Prime model.}\label{fig:clampedits}
\end{table*}

\begin{table*}[!ht]
    \tiny
    
    \centering
    \makebox[\textwidth][c]{
    \begin{tabular}{|l|l|l|l|l|l|l|l|}
    \hline
        latent\_idx & hook\_action & refuse\_id & clamp\_value & steering\_coefficient & sae\_id & sae\_release & description \\ \hline
        11766 & clamp\_refusal & 15864 & 0.05 & -500 & layer\_7/width\_16k/canonical & gemma-scope-2b-pt-res-canonical & mentions of the coronavirus pandemic and related medical terms \\ \hline
        9723 & clamp\_refusal & 15864 & 0.05 & -500 & layer\_7/width\_16k/canonical & gemma-scope-2b-pt-res-canonical & references to viral RNA and its properties \\ \hline
        4788 & clamp\_refusal & 15864 & 0.05 & -500 & layer\_7/width\_16k/canonical & gemma-scope-2b-pt-res-canonical & terms related to viral infections and their impact on health \\ \hline
        1709 & clamp\_refusal & 15864 & 0.05 & -500 & layer\_7/width\_16k/canonical & gemma-scope-2b-pt-res-canonical & terms associated with epidemiological factors in disease transmission \\ \hline
        9186 & clamp\_refusal & 15864 & 0.05 & -500 & layer\_7/width\_16k/canonical & gemma-scope-2b-pt-res-canonical & terms related to respiratory health conditions and treatments \\ \hline
        3058 & clamp\_refusal & 15864 & 0.05 & -500 & layer\_7/width\_16k/canonical & gemma-scope-2b-pt-res-canonical & references to viral infections and their biological impact \\ \hline
        13381 & clamp\_refusal & 15864 & 0.05 & -500 & layer\_7/width\_16k/canonical & gemma-scope-2b-pt-res-canonical & terms related to cybersecurity threats and vulnerabilities \\ \hline
        14050 & clamp\_refusal & 15864 & 0.05 & -500 & layer\_7/width\_16k/canonical & gemma-scope-2b-pt-res-canonical & references to mental health, particularly during the COVID-19 pandemic \\ \hline
        11078 & clamp\_refusal & 15864 & 0.05 & -500 & layer\_7/width\_16k/canonical & gemma-scope-2b-pt-res-canonical & elements related to health and safety regulations regarding public spaces \\ \hline
        16246 & clamp\_refusal & 15864 & 0.05 & -500 & layer\_7/width\_16k/canonical & gemma-scope-2b-pt-res-canonical & references to disease cases and their statistics \\ \hline
    \end{tabular}
    }
    \caption{The model edits made for the Refusal-clamp model.}\label{fig:refusalclampedits}
\end{table*}

\section{Additional Figures}

\begin{figure*}[hbt!]
    \centering
    \includegraphics[width=1\linewidth]{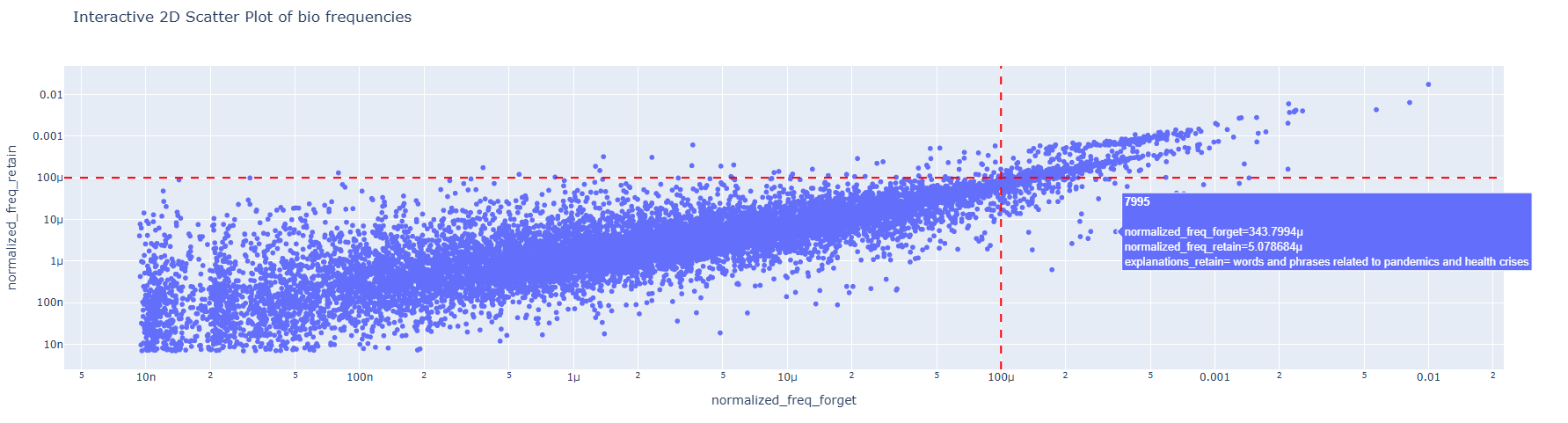}
    \caption{Frequency Forget vs Frequency Retain with a threshold marked by a horizontal dashed line. The blue box contains an example of identified harmful latent.}
    \label{fig:freq}
\end{figure*}

For the visualization of the feature frequencies, refer to Fig. \ref{fig:freq}.

\begin{figure}[H]
    \centering
    \vspace*{0pt}  
    \includegraphics[width=1\linewidth]{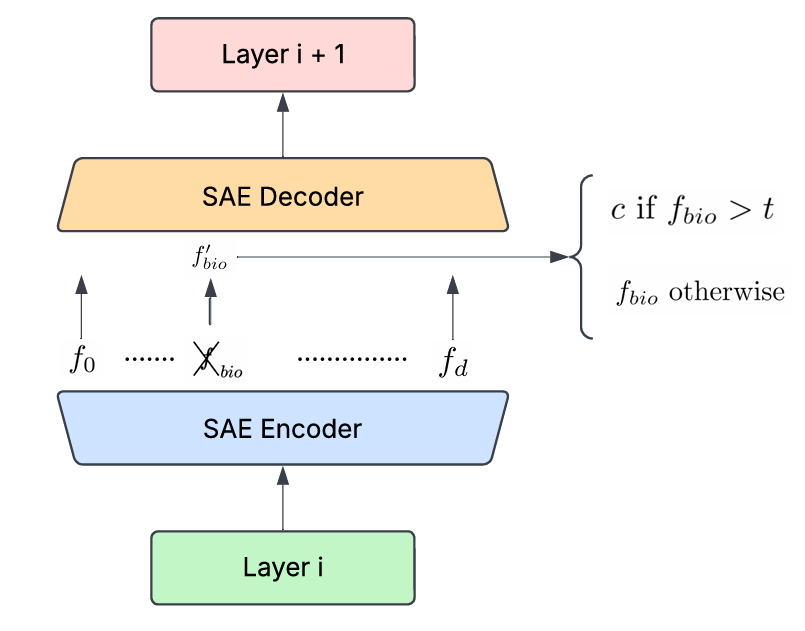}
    \caption{"Clamp Prime" algorithm}
    \label{fig:clamp}
\end{figure}

\section{Adversarial Question}\label{adversarialquestion}
Below is the randomly-selected question-answer pair from WMDP-Bio used in the adversarial evaluation. In multiple-choice format, {\tt gemma-2-2b} gets this question right regardless of how the four answers are permuted. 
\begin{quote}
\textbf{Question:} What is the role of a guanylyltransferase assay in the context of reverse genetics for studying viral capping enzymes?

\textbf{Answer:} To measure the efficiency of 5' cap addition to viral mRNAs by the capping enzyme
\end{quote}

\section{Ethical Considerations}
Unlearning techniques in language models offer a promising way to mitigate harmful biases and prevent the spread of dangerous knowledge. This is especially important as LLMs touch more lives, fall into more hands, and become more powerful each day.
Methods such as SAE-based unlearning and RMU can effectively remove access to harmful content while preserving general knowledge. However, unlearning alone does not resolve deeper issues rooted in training data; improving dataset curation is also essential to preventing bias from being embedded in models from the outset.
A key ethical risk of unlearning is potential misuse for censorship. While limiting harmful content is beneficial, these techniques could also be exploited to suppress politically sensitive information or shape narratives. As unlearning development continues, ensuring transparency in what is unlearned and implementing safeguards against ideological manipulation will be crucial to maintaining trust and accountability.

\end{document}